\documentclass{article}
\usepackage[preprint]{colm2026_conference}

\usepackage{microtype}
\usepackage{hyperref}
\usepackage{url}
\usepackage{booktabs}
\usepackage{amsfonts}
\usepackage{amsmath}
\usepackage{amssymb}
\usepackage{makecell}
\usepackage{graphicx}
\usepackage{verbatim}
\usepackage{caption}
\usepackage{lineno}

\definecolor{darkblue}{rgb}{0, 0, 0.5}
\hypersetup{colorlinks=true, citecolor=darkblue, linkcolor=darkblue, urlcolor=darkblue}

\graphicspath{{figures/}}

\title{MERIT: Modular Framework for Multimodal Misinformation Detection with Web-Grounded Reasoning}

\author{
  Mir Nafis Sharear Shopnil \\
  Fatima Fellowship \\
  \text{sharears4077@gmail} \\
  \And
  Sharad Duwal \\
  Fatima Fellowship \\
  \And
  Abhishek Tyagi \\
  University of Rochester \\
  Rochester, NY, USA \\
  \And
  Adiba Mahbub Proma \\
  University of Rochester \\
  Rochester, NY, USA
}

\begin{document}

\ifcolmsubmission
\linenumbers
\fi

\maketitle

\begin{abstract}
We present MERIT, an inference-time modular framework for multimodal misinformation detection that decomposes verification into four specialized modules: visual forensics, cross-modal alignment, retrieval-augmented claim verification, and calibrated judgment. On MMFakeBench, MERIT with GPT-4o-mini achieves 81.65\% F1, outperforming all reported zero-shot baselines including GPT-4V with MMD-Agent (74.0\% F1). A controlled same-model evaluation confirms gains stem from architectural design: MERIT achieves 6.14 points higher misinformation recall than MMD-Agent under identical model conditions, with per-class gains of +18.0 on visual distortion and +5.33 on textual distortion. Ablation studies reveal non-overlapping module specialization---removing any module disproportionately degrades its target category while leaving others intact. Test set evaluation (5,000 samples) confirms generalization within 0.21 F1 points of validation results. The framework operates with any instruction-following vision-language model and produces citation-linked rationales for human review.
\end{abstract}

\section{Introduction}

Multimodal posts combining text and images now dominate web platforms, and misinformation spreads through them far faster than fact-checkers can respond~\citep{guo-etal-2022-survey}. Multimodal misinformation, false narratives that exploit coordinated manipulation across text and images~\citep{liu2024mmfakebench}, takes many forms: false claims paired with synthetic images, authentic images repurposed with misleading captions, or genuine content placed in wrong contexts~\citep{liu2024mmfakebench, 10.1145/3696410.3714498}. These gaps motivate detection approaches that can operate without domain-specific training data and provide transparent, evidence-backed judgments.

\begin{figure*}[t]
  \centering
  \includegraphics[width=\textwidth]{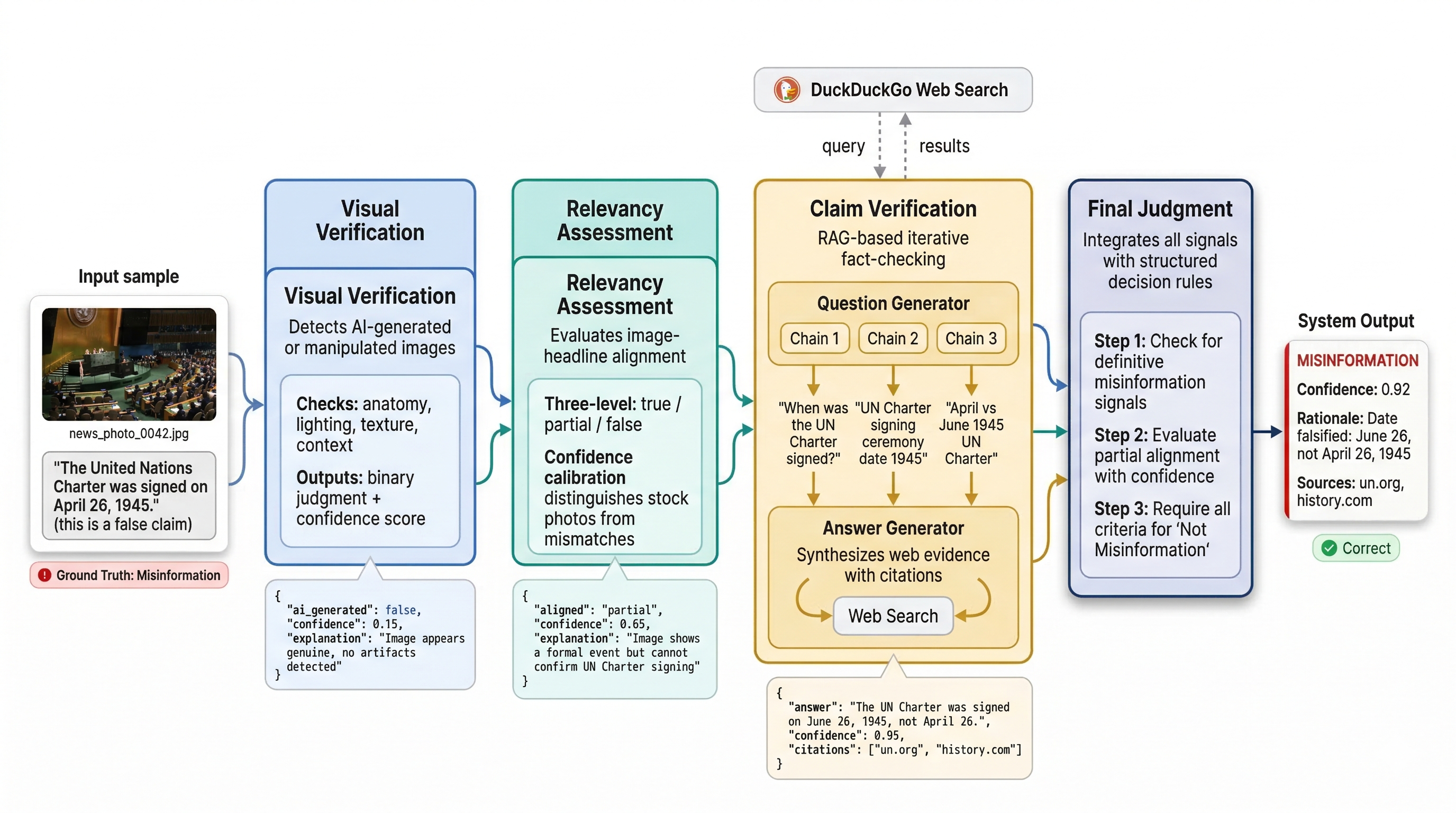}
  \caption{MERIT architecture showing the four sequential modules: Visual Verification analyzes images for AI generation and manipulation, Relevancy Assessment evaluates cross-modal alignment, Claim Verification performs retrieval-augmented fact-checking, and Final Judgment integrates all signals to produce calibrated predictions.}
  \label{fig:architecture}
\end{figure*}

Prior work has made progress on several fronts. Supervised models achieve strong benchmark performance~\citep{k2025limits}, retrieval-augmented generation mitigates parametric knowledge limits~\citep{lewis2020retrieval}, and vision-language models enable flexible multimodal reasoning~\citep{achiam2023gpt, liu2023visual}. Three gaps remain that MERIT targets. First, supervised detectors require domain-specific training data and fail to generalize across manipulation types~\citep{k2025limits, wu-etal-2024-towards}. Second, LVLMs prompted without retrieval produce inconsistent judgments and fabricate explanations when knowledge is lacking~\citep{bang-etal-2023-multitask, caramancion2023news}. Third, no prior system combines cross-modal verification, retrieval-augmented reasoning, and calibrated confidence scoring in a model-pluggable architecture.

We investigate three research questions: (RQ1)~Can a modular pipeline that decomposes multimodal verification into specialized stages outperform end-to-end vision-language model prompting? (RQ2)~What is each module's distinct contribution to detection performance? (RQ3)~Do the performance gains stem from architectural design or the underlying model?

Here, we present MERIT (Multimodal Evidence-based Retrieval for Image-Text verification), an inference-time framework with four sequential verification modules. A visual verification module detects AI-generated or manipulated images. A relevancy assessment module evaluates whether the image matches the headline using three-level alignment judgments (true, partial, false). A claim verification module generates investigative questions, retrieves web evidence via DuckDuckGo across three reasoning chains, and synthesizes citation-linked answers. A final judgment module integrates all signals using structured decision rules: to classify content as authentic, all three signals must pass—factually supported headline, aligned image, and non-AI-generated imagery. Full details appear in Section~\ref{sec:methodology}.

Our contributions are:
\begin{itemize}
    \item We propose MERIT, a modular verification framework with four specialized stages that outperforms the strongest zero-shot baselines on MMFakeBench (Table~\ref{tab:main}), produces citation-linked rationales for human review, and operates with any instruction-following vision-language model without task-specific training.
    \item We conduct a controlled same-model evaluation showing that MERIT's gains stem from architectural design rather than model capabilities 
    (Section~\ref{sec:main_results}).
    \item We analyze 249 misclassifications, identifying three systematic failure modes that inform future improvements in multimodal misinformation detection.
\end{itemize}

\section{Related work}

\subsection{Multimodal misinformation benchmarks}

Early fake news datasets focused on text-only claims or isolated image manipulations. MMFakeBench introduced mixed-source evaluation with 11,000 image-text pairs spanning three distortion categories: textual veracity distortion (false claims with real or AI images), visual veracity distortion (authentic text with manipulated images), and cross-modal consistency distortion (real text with wrong-context images)~\citep{liu2024mmfakebench}. Human annotators achieve only 56.8\% accuracy on binary classification~\citep{liu2024mmfakebench}, and LVLM baselines reach 72--74\% F1, leaving substantial room for improvement.

SNIFFER targets out-of-context detection with entity-grounded explanations~\citep{qi2024sniffer}, while MDAM3 extends coverage to audio and video alongside text and images~\citep{10.1145/3696410.3714498}. These benchmarks establish that real-world misinformation requires cross-modal reasoning rather than isolated text or image analysis.

\subsection{Vision-language models and retrieval-augmented fact-checking}

Instruction-tuned vision-language models enable multimodal reasoning over image-text pairs~\citep{liu2023visual, zhang2024somelvlm, achiam2023gpt}, but direct prompting for fact-checking yields moderate accuracy with inconsistent judgments when knowledge is lacking. Retrieval-augmented generation mitigates this by grounding language models in external documents~\citep{lewis2020retrieval, yao2023react}, and multi-agent architectures decompose verification into specialized roles~\citep{ma2025local, liu2024can}. Systems augmenting LVLMs with external knowledge improve reliability~\citep{xuan2024lemma, khaliq2024ragar, wan2024dell}, but most focus on text-only claims and none combine explicit visual forensics with cross-modal alignment and calibrated confidence aggregation. MERIT extends tool-augmented decomposition to multimodal settings with a model-pluggable design where any instruction-following LVLM can serve as the reasoning engine.

\subsection{Visual forensics and cross-modal consistency}

Detecting AI-generated and manipulated images requires complementary approaches. Surveys identify four detection families: statistical features analyzing noise patterns and compression artifacts, model likelihood methods probing generator fingerprints, supervised classifiers trained on synthetic datasets, and provenance techniques such as watermarks and C2PA standards~\citep{10.1145/3589335.3641256}.

Cross-modal consistency verification detects when authentic images are paired with misleading captions. \citet{10.1145/3696410.3714498} combine internal visual detectors with external web signals, showing that multi-source explanations improve human understanding compared to binary labels. Visual veracity and cross-modal consistency are related but distinct problems. MERIT addresses both through separate modules: visual verification for synthetic content detection and relevancy assessment for image-text alignment.

\subsection{LLM reliability, calibration, and limitations}

Unchecked LLM judgments risk bias and inconsistency, with studies showing systematic divergence from journalist baselines~\citep{yang2025accuracy} and only 61--72\% accuracy on fact-checked claims without retrieval~\citep{caramancion2023news}. Frameworks addressing reliability emphasize calibrated confidence scores and abstention triggers~\citep{pelrine2023towards, 10.1145/3696410.3714640}. MERIT mitigates these concerns through module-level confidence scores that feed into structured decision rules, and citation-linked rationales that expose the evidence driving each decision.

\section{Methodology}
\label{sec:methodology}

\subsection{Overview}

MERIT addresses multimodal misinformation detection as a structured reasoning problem. Given an image $I$ and text headline $H$, the system predicts whether the pair is misinformation or not, producing a binary classification: Misinformation or Not Misinformation. Unlike supervised models that learn detection patterns from labeled training data, MERIT decomposes verification into four sequential modules that each address a distinct aspect of the detection task.

The pipeline operates as follows. The Visual Verification Module analyzes image $I$ for signs of AI generation or manipulation. The Relevancy Assessment Module evaluates whether image and text align semantically. The Claim Verification Module generates investigative questions about headline $H$, retrieves evidence via web search following retrieval-augmented generation principles~\citep{lewis2020retrieval}, and synthesizes answers with citations. The Final Judgment Module integrates all signals to produce a calibrated binary decision. 

Figure~\ref{fig:architecture} illustrates this architecture. Formally, MERIT computes the probability of misinformation through modular composition:

\begin{equation}
P(\text{Misinformation} \mid I, H) = f_{\text{judge}}(v, a, r)
\end{equation}
where:
\begin{align}
v &= f_{\text{visual}}(I) \quad \text{(visual verification)} \\
a &= f_{\text{align}}(I, H) \quad \text{(relevancy assessment)} \\
r &= f_{\text{RAG}}(H, \mathcal{W}) \quad \text{(claim verification)}
\end{align}

Here, $f_{\text{visual}}: \mathcal{I} \rightarrow \{0,1\} \times [0,1]$ detects AI-generated or manipulated images, outputting a binary indicator and confidence score. $f_{\text{align}}: \mathcal{I} \times \mathcal{H} \rightarrow \{\text{true, partial, false}\} \times [0,1]$ evaluates cross-modal consistency, producing alignment level and confidence. $f_{\text{RAG}}: \mathcal{H} \times \mathcal{W} \rightarrow \mathcal{Q} \times \mathcal{A}$ generates investigative questions $\mathcal{Q}$ about headline $H$, retrieves evidence from the web $\mathcal{W}$, and synthesizes citation-linked answers $\mathcal{A}$. Finally, $f_{\text{judge}}$ aggregates signals $(v, a, r)$ through a three-step decision process: first checking for definitive misinformation signals (false claims, AI-generated images, or mismatched content), then evaluating ambiguous partial-alignment cases using confidence thresholds, and finally requiring all three signals to pass for a ``Not Misinformation'' classification. The complete decision logic is specified in Appendix~\ref{sec:judge_prompt}. The pipeline processes each module sequentially, with retrieval-augmented claim verification querying DuckDuckGo web search following RAG principles~\citep{lewis2020retrieval}.
 
We implement all modules using GPT-4o-mini~\citep{hurst2024gpt}, accessed via the OpenAI API with temperature set to 0 for deterministic outputs.

\subsection{Visual verification module}

The visual verification module detects AI-generated or manipulated images through structured prompting of the vision-language model. The module receives a base64-encoded image as input and instructs GPT-4o-mini to analyze the image for characteristic artifacts that indicate synthetic or altered content.

The prompt directs the model to examine both technical artifacts (warped hands, inconsistent lighting, impossible anatomy) and contextual anomalies (surreal object combinations, dreamlike elements, impossible scenarios). The model outputs structured JSON containing a binary AI-generation judgment, confidence score calibrated between 0 and 1, explanation of observed patterns, and a list of specific anomalies detected. See Appendix for complete prompt text.

This approach differs from traditional deepfake detectors that analyze statistical properties or generator fingerprints~\citep{10.1145/3589335.3641256}. Instead, MERIT leverages the vision-language model's semantic understanding to identify both technical imperfections and contextual anomalies, such as impossible object combinations, physics violations, or scenes that appear 
plausible at first glance but contain logically inconsistent elements that may not be captured by statistical forensic detectors focused on pixel-level artifacts ~\citep{10.1145/3589335.3641256}.
\subsection{Relevancy assessment module}

The relevancy assessment module evaluates whether the image depicts the specific subject, event, or context described in the headline. This addresses out-of-context misinformation where authentic images are repurposed with misleading captions.

The module produces a categorical alignment judgment $a \in \{\text{true}, \text{partial}, \text{false}\}$ paired with a confidence score $c \in [0,1]$. True alignment ($a = \text{true}$) indicates the image depicts the specific subject or event in the headline. Partial alignment ($a = \text{partial}$) indicates related content without confirmation of specifics. False alignment ($a = \text{false}$) indicates a different subject or direct contradiction.

The prompt calibrates confidence scores to distinguish legitimate partial alignment ($\geq$0.7, right subject with incomplete details) from deceptive mismatches ($<$0.7, superficial similarity). See Appendix for complete prompt.

\subsection{Claim verification module}

The claim verification module performs retrieval-augmented fact-checking through a two-stage process: question generation followed by answer synthesis. The module generates investigative queries across three sequential reasoning chains, retrieves 
web evidence for each query, and synthesizes citation-linked answers. This iterative design enables deeper investigation than single-pass retrieval. Complete prompt templates and example outputs appear in Appendix~\ref{sec:claim_prompts}.


The question generator produces investigative queries in three sequential chains, each generating three questions ($k$=3) with duplicate detection across chains. Chain~1 verifies core claims (``Did [event] happen?''). Chain~2 checks specific details informed by Chain~1 findings. Chain~3 resolves remaining ambiguities. Concrete example outputs appear in the Appendix.

For each question, the system queries DuckDuckGo and retrieves up to 5 results. The answer synthesis prompt receives titles, URLs, and snippets, then produces a concise answer (2--5 sentences) with explicit source citations. When sources conflict, the model summarizes differing perspectives and cites each separately. The output includes a calibrated confidence score; low scores signal sparse or contradictory evidence, which the judgment module uses to avoid overconfident classification. Complete prompts appear in the Appendix.

\section{Experimental setup}

We evaluate MERIT on MMFakeBench~\citep{liu2024mmfakebench}, containing 11,000 image-text pairs across three distortion categories (textual, visual, cross-modal consistency) plus authentic news from VisualNews. The benchmark splits into validation (1,000 samples) and test (10,000 samples) sets with approximately 70\% misinformation and 30\% authentic content. We conduct primary evaluation on the validation set and confirm generalization on 5,000 stratified test samples (seed 42).





We follow the MMFakeBench binary classification protocol~\citep{liu2024mmfakebench} with ``Misinformation'' as the positive class. We report F1 score as the primary metric for comparability with prior work, alongside accuracy, precision, and per-class recall. We additionally report balanced accuracy ($\frac{\text{Sensitivity} + \text{Specificity}}{2}$) in ablation studies to expose configurations that exploit class imbalance rather than performing genuine verification~\citep{brodersen2010balanced}, and assess threshold robustness using PR-AUC and ROC-AUC.


MERIT uses GPT-4o-mini (gpt-4o-mini-2024-07-18) via the OpenAI API with temperature 0 for deterministic outputs. Visual verification and relevancy modules use vision--language prompting with base64-encoded images. For claim verification, we use DuckDuckGo's free API with 3 question chains $\times$ 3 questions per sample, retrieving up to 5 results per question. Cost and reliability details appear in the Appendix.




\subsection{Baselines}

We compare MERIT against all baselines reported in \citet{liu2024mmfakebench}. The primary baselines are GPT-4V with standard prompting (72.3\% F1, 75.6\% accuracy on validation) and GPT-4V with the MMD-Agent framework (74.0\% F1, 76.8\% accuracy). MMD-Agent decomposes the detection task hierarchically into textual veracity checking, visual veracity checking, and cross-modal consistency reasoning, then integrates reasoning with Wikipedia-based external knowledge retrieval. Additional baselines include open-source LVLMs ranging from 7B to 34B parameters (LLaVA-1.6, InstructBLIP, BLIP2, VILA) with F1 scores between 7.9\% and 67.2\%, as well as traditional detection methods. Complete baseline results are available in Table~6 of \citet{liu2024mmfakebench}.

\paragraph{Controlled same-model baseline.}
To isolate MERIT's architectural contribution from underlying model differences, a confounder present when comparing GPT-4o-mini against GPT-4V baselines, we re-implement MMD-Agent using GPT-4o-mini, the same model powering MERIT. We follow the original three-stage pipeline (textual veracity check via Wikipedia retrieval, visual veracity check, cross-modal consistency reasoning) as described in \citet{liu2024mmfakebench}, using identical temperature settings (0.0) and evaluation protocols. 

\section{Results and analysis}

We present comprehensive evaluation results on the MMFakeBench validation set (1,000 samples), including main performance comparisons, ablation studies, and per-class breakdowns. All experiments use identical stratified sampling to ensure fair comparisons across configurations.

\subsection{Performance comparison}
\label{sec:main_results}

Table~\ref{tab:main} presents our main results compared to baseline methods from \citet{liu2024mmfakebench}. MERIT achieves 81.65\% F1 on the validation set, outperforming GPT-4V with standard prompting (72.3\% F1) by 9.35 percentage points and GPT-4V with MMD-Agent (74.0\% F1) by 7.65 points. This represents the strongest reported performance on MMFakeBench validation using vision-language models.

While overall accuracy (75.1\%) is comparable to baseline systems (75.6--76.8\%), MERIT's advantage lies in misinformation recall: 79.14\% vs.\ 73.00\% (+6.14 points), catching substantially more actual misinformation. The per-class breakdown (Table~\ref{tab:same_model_perclass}) shows this gain concentrates in visual distortion (+18.0) and textual distortion (+5.33), the categories where MERIT's specialized modules differ most from MMD-Agent's approach.

\paragraph{Same-model controlled comparison.}
Table~\ref{tab:main} includes our re-implementation of MMD-Agent using GPT-4o-mini, enabling a controlled comparison that isolates framework contributions from model differences. Under identical model conditions, MERIT outperforms MMD-Agent by 2.85 F1 points (81.65\% vs.\ 78.80\%) and 2.60 accuracy points (75.1\% vs.\ 72.50\%). Misinformation recall improves by 6.14 points (79.14\% vs.\ 73.00\%), confirming that the performance gains stem from architectural design rather than the underlying language model.

\paragraph{Framework portability.}
The re-implementation exposes portability issues in MMD-Agent: its visual stage refused image analysis in 93.1\% of samples, and Wikipedia retrieval returned \texttt{INSUFFICIENT} for 54.6\% of queries. Table~\ref{tab:same_model_perclass} shows MERIT's gains are largest for visual distortion (+18.0 points) and textual distortion (+5.33 points). MMD-Agent achieves 5.66 points higher specificity on authentic content, as its visual refusal pattern incidentally preserves some real news at the cost of missing genuine misinformation.

\begin{table*}[t]
  \caption{Main results on MMFakeBench validation set}
  \label{tab:main}
  \centering
  \begin{tabular}{llccccc}
    \toprule
    Method & Model & F1 & Acc & Prec & Sens & Spec \\
    \midrule
    GPT-4V Standard$^\dagger$ & GPT-4V & 72.3 & 75.6 & 72.1 & 72.8 & --- \\
    GPT-4V MMD-Agent$^\dagger$ & GPT-4V & 74.0 & 76.8 & 73.4 & 75.5 & --- \\
    \midrule
    MMD-Agent$^*$ & GPT-4o-mini & 78.80 & 72.50 & 85.59 & 73.00 & 71.33 \\
    MERIT & GPT-4o-mini & \textbf{81.65} & \textbf{75.1} & 84.3 & \textbf{79.14} & 65.67 \\
    \bottomrule
  \end{tabular}

  \vspace{2mm}
  \small
  $\dagger$ Baseline results from \citet{liu2024mmfakebench}; specificity not reported. \\
\end{table*}
\vspace{-2mm}

Test set evaluation on 5,000 samples\footnote{Stratified random sample (50\%) preserving the original 70-30 class distribution of the full 10,000-sample test set, selected once with seed 42 for computational efficiency.} yields 81.44\% F1 and 75.08\% accuracy, confirming that performance generalizes beyond the validation set with less than 0.3 point variation in both metrics.

\begin{figure}[t]
  \centering
  \includegraphics[width=0.80\columnwidth,trim={4mm 2mm 4mm 2mm},clip]{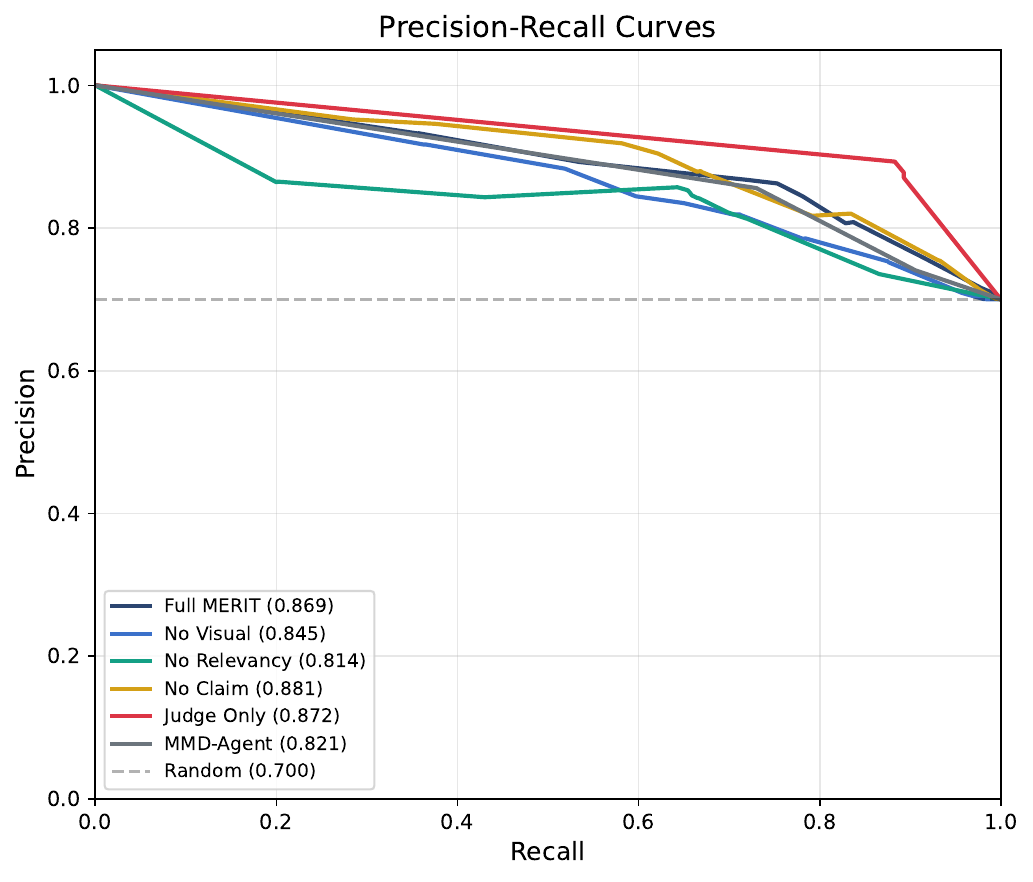}
  \caption{Precision--recall curves on the MMFakeBench validation set. Because misinformation prevalence is approximately 70\%, the no-skill baseline is 0.700. Full MERIT improves PR-AUC over same-model MMD-Agent under class imbalance.}
  \label{fig:pr_curve}
\end{figure}

To assess threshold robustness under MMFakeBench's 70:30 class skew, we also compare precision--recall and ROC curves (Figure~\ref{fig:pr_curve}; Appendix Figure~\ref{fig:roc_curve}). PR-AUC is especially informative in this setting because the no-skill baseline equals the misinformation prevalence (0.700). Full MERIT achieves 0.869 PR-AUC and 0.777 ROC-AUC, improving over same-model MMD-Agent at 0.821 and 0.723, respectively. Some ablations give slightly higher AUCs, indicating that threshold-free ranking alone does not fully capture usable verification behavior. In particular, Table~\ref{tab:ablation} shows that strong ranking scores do not guarantee a practical operating point: Judge Only performs poorly on authentic content under its default labeling behavior despite favorable AUC values. We therefore treat PR-AUC and ROC-AUC as complementary evidence of ranking robustness, while relying on F1, balanced accuracy, and per-class recall for operational comparison.

\subsection{Ablation studies}

Table~\ref{tab:ablation} demonstrates each module's contribution through systematic ablations. The judge-only configuration illustrates why F1 alone can mislead under class imbalance: it achieves the highest F1 (82.74) by classifying nearly everything as misinformation (100\% sensitivity, 2.67\% specificity), 
but its balanced accuracy collapses to 51.34\%, near random chance. MERIT achieves 72.41\% balanced accuracy, a 21-point gap confirming genuine cross-class discrimination.

Removing visual verification drops overall accuracy by 6 percentage points (69.1\% vs 75.1\%) and F1 by 5.18 points, with fake recall decreasing from 79.14\% to 71.71\%. More critically, Table~\ref{tab:perclass} reveals this primarily impacts visual distortion detection, dropping from 92\% to 56\% accuracy, a 36-point degradation. This validates that explicit visual analysis is essential for detecting AI-generated and manipulated imagery.

Removing claim verification achieves higher real recall (76.33\%) but lower fake recall (71.43\%) and reduced F1 (78.68\%), with textual distortion dropping from 84.33\% to 68.67\%.
 
Removing relevancy assessment drops F1 by 4.70 points, with the sharpest impact on cross-modal mismatch detection (69.67\%$\to$56.0\%, a 13.67-point loss). Visual distortion actually rises to 96.0\% without the relevancy module, since its signal no longer introduces noise for clear visual forgery cases.
 
This non-overlapping specialization justifies the four-module architecture: each module's removal disproportionately degrades its target category while leaving others largely intact.

\begin{table*}[t]
  \caption{Ablation study results (validation set)}
  \label{tab:ablation}
  \centering
  \begin{tabular}{lcccccccr}
    \toprule
    Configuration & F1 & Bal.\ Acc & Acc & Sens & Spec & FP Rate & $\Delta$ F1 \\
    \midrule
    Full MERIT & \textbf{81.65} & \textbf{72.41} & \textbf{75.1} & 79.14 & 65.67 & 34.3 & --- \\
    No Visual Verification & 76.47 & 67.36 & 69.1 & 71.71 & 63.0 & 37.0 & $-$5.18 \\
    No Relevancy Assessment & 76.95 & 67.36 & 69.5 & 72.71 & 62.0 & 38.0 & $-$4.70 \\
    No Claim Verification & 78.68 & 73.88 & 72.9 & 71.43 & 76.33 & 23.7 & $-$2.97 \\
    Judge Only & 82.74 & 51.34 & 70.8 & \textbf{100.0} & 2.67 & 97.3 & $+$1.09 \\
    \bottomrule
  \end{tabular}

  \vspace{2mm}
  \small
  $\Delta$ F1 = Change in F1 score relative to full MERIT. \\
  FP Rate = False Positive Rate on authentic content. \\
  Sens = Recall (Misinformation); Spec = Recall (Not Misinformation).
\end{table*}
\vspace{-2mm}

\subsection{Per-class performance analysis}

Table~\ref{tab:perclass} confirms non-overlapping module specialization. Each module's removal disproportionately degrades its target category: visual verification owns visual distortion (92\%$\to$56\%), relevancy assessment owns cross-modal mismatch (69.67\%$\to$56\%), and claim verification owns textual distortion (84.33\%$\to$68.67\%). Cross-modal mismatches remain hardest (69.67\%), as out-of-context posts pair genuine imagery with partly true captions that blur the boundary between illustrative reporting and deliberate miscontextualization.

\begin{table}[t]
  \caption{Detection accuracy by misinformation type}
  \label{tab:perclass}
  \centering
  \begin{tabular}{lccccc}
    \toprule
    Type & Full & \makecell{No\\Visual} & \makecell{No\\Relev.} & \makecell{No\\RAG} & \makecell{Judge\\Only} \\
    \midrule
    Visual distortion & 92.0 & 56.0 & 96.0 & 92.0 & 100.0 \\
    Textual distortion & 84.33 & 80.67 & 81.67 & 68.67 & 100.0 \\
    Cross-modal mismatch & 69.67 & 68.0 & 56.0 & 67.33 & 100.0 \\
    Authentic content & 65.67 & 63.0 & 62.0 & 76.33 & 2.67 \\
    \bottomrule
  \end{tabular}

  \vspace{2mm}
  \small
  Visual module critical for AI images; RAG critical for textual lies.
\end{table}
\vspace{-2mm}

\begin{table}[t]
  \caption{Same-model per-class accuracy (\%): MERIT vs.\ 
MMD-Agent, both GPT-4o-mini.}
  \label{tab:same_model_perclass}
  \centering
  \begin{tabular}{lccc}
    \toprule
    Misinformation Type & MERIT & MMD-Agent & $\Delta$ \\
    \midrule
    Visual distortion & 92.0 & 74.0 & +18.0 \\
    Textual distortion & 84.33 & 79.0 & +5.33 \\
    Cross-modal mismatch & 69.67 & 66.67 & +3.0 \\
    Authentic content & 65.67 & 71.33 & $-$5.66 \\
    \midrule
    Overall F1 & 81.65 & 78.80 & +2.85 \\
    \bottomrule
  \end{tabular}
\end{table}
\vspace{-2mm}

\subsection{Error analysis}

We analyzed all 249 errors from the 1,000-sample validation run to identify systematic failure patterns. False positives (103 errors, 34.3\% of real news) occur primarily due to weak image-text alignment (61.2\% of false positives). Legitimate news articles frequently use generic stock photos or illustrative images rather than literal depictions of events. For example, an authentic VisualNews article about a beached whale was flagged as fake because the image showed a whale but could not confirm the specific location mentioned in the headline. The system incorrectly interpreted partial alignment as evidence of manipulation.

Unusual but true events also trigger false alarms when web search returns limited results for niche topics. A genuine sports article about Northwestern's Gator Bowl victory was flagged as fake simply because the specific game had minimal online coverage, demonstrating that absence of search results does not equal evidence of falsehood but that can be tricky to detect.

MMD-Agent's errors stem from visual module collapse (93.1\% refusal rate on GPT-4o-mini) and sparse Wikipedia coverage (54.6\% insufficient), while MERIT's errors concentrate on cross-modal alignment for ambiguous cases. Detailed analysis appears in the Appendix.

\section{Discussion and future work}
 



MERIT addresses the three research questions posed in Section~1: modular decomposition outperforms end-to-end prompting (RQ1, Table~\ref{tab:main}), each module contributes to a distinct detection category with minimal overlap (RQ2, Tables~\ref{tab:ablation}--\ref{tab:perclass}), and same-model comparison confirms that the gains arise from architectural design rather than model choice (RQ3, Table~\ref{tab:same_model_perclass}). These results suggest that multi-modal misinformation detection benefits from decomposing verification into explicit subproblems, visual authenticity, cross-modal alignment, and evidence-grounded claim checking, rather than asking a single prompt to resolve all failure modes at once, consistent with prior work showing that multi-modal misinformation requires genuine cross-modal reasoning and that retrieval helps mitigate model knowledge limits~\citep{liu2024mmfakebench, 10.1145/3696410.3714498, lewis2020retrieval}. 

The framework is model-pluggable and produces citation-linked rationales for transparent auditing~\citep{dougrez2024assessing}, although future work should add a citation-validation step because LLM judgments can remain unreliable when evidence is sparse or ambiguous~\citep{yang2025accuracy, caramancion2023news}. Because MERIT is search-dependent, very recent or niche events may be weakly indexed, which likely contributes to some of the hard cases in our error analysis. MMFakeBench's 70:30 class skew also means that naive all-fake strategies can appear competitive on accuracy, so our decision rules require positive evidence from each module; despite being a single-benchmark evaluation, MMFakeBench remains an appropriate testbed because it combines textual distortion, visual distortion, cross-modal mismatch, and authentic news in a mixed-source setting that explicitly stresses cross-modal verification~\citep{liu2024mmfakebench, 10.1145/3696410.3714498}. 

In practice, MERIT is better suited to batch verification or human-assisted review than real-time moderation, given its reliance on commercial APIs, processing cost, and nontrivial false-positive rate, but the controlled evaluation shows that its modular design transfers more robustly across models and allows individual components to be updated without retraining the full system; future work should therefore focus on multilingual extension, threshold tuning to control the sensitivity--specificity trade-off, and tighter integration of citation-linked outputs into human review workflows~\citep{pelrine2023towards, 10.1145/3696410.3714640}.

\section{Conclusion}

We presented MERIT, a modular framework that decomposes multimodal misinformation detection into four specialized verification stages. Controlled evaluation on MMFakeBench confirms that this decomposition outperforms zero-shot baselines, with same-model comparison confirming gains stem from architectural design. Ablation studies validate that each module addresses a distinct failure mode with minimal overlap. Error analysis identifies systematic failure patterns that inform the future directions discussed above.

\section*{Ethics statement}

Automated misinformation detection systems raise ethical concerns, requiring careful deployment considerations. Vision-language models inherit training data biases that may affect detection accuracy across demographic groups or political viewpoints. The dual-use nature of detection systems means the same techniques can be reverse-engineered to identify weaknesses in the system and bypass them. Researchers should take these considerations into account while designing systems. Future work should focus on evaluating biases in the system and ensuring safe deployment of the models.

We position MERIT as a tool for fact-checkers, platform moderators, and end-users. It is not a replacement for media literacy education and appropriate policies. To ensure responsible use of such systems, we recommend ongoing dialog among researchers, platforms, policymakers, and civil society.

\section*{LLM usage disclosure}
GPT-4o-mini serves as the inference engine in the MERIT framework; this constitutes the core research contribution. LLM-based writing assistants were used to help draft and edit portions of the manuscript text. All content was reviewed and verified by the authors.

\bibliography{references}
\bibliographystyle{colm2026_conference}

\appendix
\section{System prompts}

This appendix documents the complete prompt templates used in each MERIT module. All prompts instruct GPT-4o-mini to output strict JSON for downstream processing.

\subsection{Visual verification module}

The visual verification module uses a two-part prompt structure to detect AI-generated or manipulated images.

\subsubsection{System prompt}

\begin{verbatim}
You are an AI image detection expert. Analyze
images for signs of AI generation or manipulation.
Respond with strict JSON only.
\end{verbatim}

\subsubsection{User instruction}

\begin{verbatim}
Task: Examine this image for signs of AI generation
or manipulation.

Detection criteria:

Technical artifacts:
- Warped hands, extra fingers, impossible anatomy
- Nonsensical text, garbled signs
- Inconsistent lighting, impossible shadows
- Unnatural textures, blending errors

Contextual anomalies:
- Surreal object combinations
  (e.g., clown in bathroom mirror)
- Impossible scenarios or physics violations
- Dreamlike or fantastical elements in otherwise
  normal scenes
- Objects that don't belong in the context

Be suspicious when:
- Technically perfect BUT contextually bizarre
- Minor details don't make sense
- Scene feels 'off' even without obvious artifacts

Confidence calibration:
- 0.8-1.0: Clear technical artifacts OR
           impossible context
- 0.6-0.8: Strong evidence, multiple anomalies
- 0.4-0.6: Moderate suspicion, some anomalies
- 0.2-0.4: Minor concerns, could be unusual
           real photo
- 0.0-0.2: Appears genuine

Prioritize contextual impossibility as much as
technical quality.

Output JSON:
{
  "ai_generated": boolean,
  "confidence": 0.0-1.0,
  "explanation": "What you observed",
  "anomalies": ["specific issues"]
}
\end{verbatim}

The prompt directs the model to examine both technical artifacts (warped anatomy, inconsistent lighting) and contextual anomalies (surreal combinations, impossible scenarios). Confidence thresholds guide calibrated scoring where higher values indicate stronger evidence of AI generation.

\subsection{Relevancy assessment module}

The relevancy module evaluates image-text alignment through three-level classification to detect out-of-context misinformation.

\subsubsection{System prompt}

\begin{verbatim}
You are an image-headline relevancy assessor.
Evaluate if the image depicts the specific subject
and context described in the headline.
Respond with strict JSON only.
\end{verbatim}

\subsubsection{User instruction template}

\begin{verbatim}
Headline: {headline}

Task: Does the image show the specific subject/event
from the headline?

Classification:
- aligned=true: Image clearly depicts the specific
  subject/event
- aligned=partial: Image shows related content but
  lacks confirmation of specifics
- aligned=false: Image shows different subject/event

Confidence calibration (CRITICAL):
- 0.9-1.0: Can identify specific people/places/events
  mentioned in headline
- 0.7-0.9: Shows correct general context but cannot
  confirm specific details
- 0.5-0.7: Shows related content but connection is
  weak or ambiguous
- 0.3-0.5: Superficial similarity only, likely
  wrong subject
- 0.0-0.3: No meaningful connection

For partial alignment:
- High confidence (0.7+): Right subject, details
  not fully visible
- Low confidence (<0.7): Possibly wrong subject,
  superficial match

Output JSON:
{
  "aligned": "true" | "partial" | "false",
  "confidence": 0.0-1.0,
  "explanation": "What you observed and why this
                  confidence"
}
\end{verbatim}

The prompt explicitly defines confidence ranges to distinguish legitimate partial alignment, common in journalistic practices where stock photos illustrate stories, from deceptive cross-modal mismatches where images depict different subjects than captions claim.

\subsection{Claim verification: question generation}

The question generator produces investigative queries in three sequential chains, enabling iterative investigation where each chain builds on findings from previous questions.

\subsubsection{System prompt}

\begin{verbatim}
You are an investigative assistant. Generate search
queries to verify if an image-headline pairing is
authentic or misleading. Focus on verifying both
the headline's claims AND whether the image matches.
Respond with strict JSON only.
\end{verbatim}

\subsubsection{User instruction template}

\begin{verbatim}
Task: Generate {k} Google-style search queries to
verify this headline.

Headline: {headline}

Already asked:
{prior_questions}

Recent answers:
{answered_qa_pairs}

Query strategy:
- First query: Verify core claim exists
  ("Did [event] happen?", "Is [fact] true?")
- Follow-up queries: Check specific details, dates,
  people involved
- Use concrete terms: names, places, dates,
  specific events
- Keep queries short (4-8 words)

Generate exactly {k} NEW queries (avoid duplicates).

Output as JSON array only:
["query 1", "query 2", ...]
\end{verbatim}

The template includes prior questions and recent answers to enable adaptive investigation. The system generates three questions per chain ($k$=3), with explicit duplicate detection preventing redundant searches across the three chains. Each chain explores different verification angles: Chain 1 addresses direct fact-checking, Chain 2 generates context questions informed by Chain 1 findings, and Chain 3 asks follow-up questions to resolve ambiguities from earlier chains.

\subsection{Claim verification: answer synthesis}
\label{sec:claim_prompts}

The answer generator synthesizes citation-linked responses from web search results, grounding claims in external evidence rather than relying on parametric knowledge.

\subsubsection{System prompt}

\begin{verbatim}
You are a careful fact-checking assistant. Using
the provided web snippets, answer the user's
question concisely and cite sources. If sources
disagree, summarize the differing views and cite
each. Respond with strict JSON only.
\end{verbatim}

\subsubsection{User instruction template}

\begin{verbatim}
Question: {question}

Sources:
[1] {title_1}
URL: {url_1}
Snippet: {description_1}

[2] {title_2}
URL: {url_2}
Snippet: {description_2}

[... up to 5 sources ...]

Instructions: Produce strict JSON with keys:
  answer: short textual answer (2-5 sentences)
  citations: array of objects {url, title} for
            the sources you used
  confidence: number in [0,1]
  rationale: one or two sentences on how you
            arrived at the answer
\end{verbatim}

The prompt instructs the model to synthesize concise answers from provided search result snippets, include calibrated confidence scores between 0 and 1, and cite sources explicitly by URL and title. When sources present conflicting information, the model summarizes differing perspectives and cites each source, enabling transparent assessment of evidence quality. This citation-grounded approach ensures answers remain traceable to source material for audit purposes.

\subsection{Final judgment module}
\label{sec:judge_prompt}

The judgment module integrates all verification signals using structured decision rules designed to prevent naive classification strategies that exploit class imbalance.

\subsubsection{System prompt}

\begin{verbatim}
You are a misinformation detector. Evaluate
image-headline pairings using multiple signals.

Input signals:
- relevancy: {aligned, confidence, explanation}
- visual_veracity: {ai_generated, confidence,
                    anomalies}
- qa_analysis: Verification of headline claims

Decision logic:

STEP 1 - Definitive misinformation (ANY of these):
- Headline verifiably false (Q/A contradicts)
- Image is AI-generated (ai_generated=true,
  confidence>0.6)
- Image completely wrong (aligned=false)
-> Classify as Misinformation

STEP 2 - Evaluate partial alignment cases:
When aligned=partial, USE confidence to distinguish:

High confidence partial (>=0.7):
- Interpretation: Right subject, incomplete
  details visible
- If headline true AND image genuine
  -> Not Misinformation

Low confidence partial (<0.7):
- Interpretation: Possibly wrong subject,
  superficial match
- If headline true AND image genuine
  -> Misinformation (likely mismatch)

STEP 3 - Verify genuine content (ALL required):
- Headline accurate (Q/A supports)
- Image genuinely relates (aligned=true OR
  partial with confidence>=0.7)
- Image authentic (ai_generated=false)
-> Not Misinformation

Return JSON:
{
  "label": "Misinformation" | "Not Misinformation",
  "confidence": 0.0-1.0,
  "rationale": "Which signals were decisive",
  "key_factors": ["signal: value"]
}
\end{verbatim}

\subsubsection{User message template}

\begin{verbatim}
You are given the following analysis JSON for a
headline+image pair. Make a final misinformation
judgment.

Analysis JSON (compact):
{
  "headline": "{headline_text}",
  "image_path": "{path}",
  "relevancy": {aligned, confidence, explanation},
  "visual_veracity": {ai_generated, confidence,
                      anomalies},
  "best_qa_per_chain": [
    {
      "question": "{q1}",
      "answer": "{a1}",
      "confidence": 0.X,
      "citations_count": N
    },
    ...
  ]
}

Respond ONLY with JSON:
{"label":..., "confidence":..., "rationale":...,
 "key_factors":[...]}
\end{verbatim}

The judgment prompt encodes explicit decision rules to prevent the model from defaulting to majority-class predictions. Classification as Not Misinformation requires satisfying all three criteria simultaneously: factual headline accuracy verified through question-answer evidence, genuine image-text alignment (true or high-confidence partial), and authentic non-AI-generated imagery. This asymmetric decision structure forces comprehensive verification before labeling content as legitimate, addressing the class imbalance problem where naive ``classify everything as fake'' strategies achieve high accuracy but unacceptable false positive rates.

\subsection{Implementation notes}

All prompts use temperature 0 for deterministic outputs, enabling exact replication across runs. The visual verification and relevancy modules receive base64-encoded images via GPT-4o-mini's vision capabilities, while the claim verification modules operate on text only. JSON output formatting is enforced through explicit prompt instructions rather than OpenAI function calling to maintain compatibility across model versions and facilitate future integration with alternative vision-language models. The structured output format enables programmatic parsing and integration into content moderation pipelines requiring audit trails.

\section{Cost and reliability details}

Evaluation on the 1{,}000-sample validation set used 103M prompt tokens and 1.4M completion tokens. At \$0.15/M (prompt) and \$0.60/M (completion), cost is $\approx$\$16.29 per 1{,}000 samples; using GPT-4V instead would be $\approx$\$536.45---33$\times$ higher. Reliability safeguards include exponential-backoff retries (up to 2), 35\,s per-query timeouts, a 1.8\,s rate limit, and result caching.

\section{Detailed failure mode comparison}
\label{sec:failure_modes}

The same-model MMD-Agent baseline exhibits different error patterns from MERIT. Its dominant failure is visual verification collapse: GPT-4o-mini refuses image analysis in 93.1\% of samples and defaults to 
``authentic'' labels, missing AI-generated images that MERIT detects at 92.0\% accuracy. Its second failure source is sparse Wikipedia coverage---54.6\% of textual veracity queries returned \texttt{INSUFFICIENT} evidence, compared to MERIT's DuckDuckGo-based retrieval which accesses a broader web corpus.

\section{Question generation example}

For the headline ``The United Nations Charter was signed on April 26, 1945,'' the three chains produce:

\textbf{Chain 1} (direct fact-checking): ``When was the UN Charter signed?'', ``UN Charter signing date 1945'', ``April 26 1945 UN Charter.''

\textbf{Chain 2} (informed by Chain 1 evidence that the actual date was June 26): ``UN Charter signing ceremony San Francisco'', ``June 26 1945 UN signing details'', ``difference April June 1945 UN.''

\textbf{Chain 3} (resolving gaps): ``UN Charter signatories 1945'', ``UN Charter ratification date'', 
``UN Charter vs UN Declaration.''

\section{Threshold-robustness curves}

\begin{figure}[t]
  \centering
  \includegraphics[width=0.82\columnwidth,trim={4mm 2mm 4mm 2mm},clip]{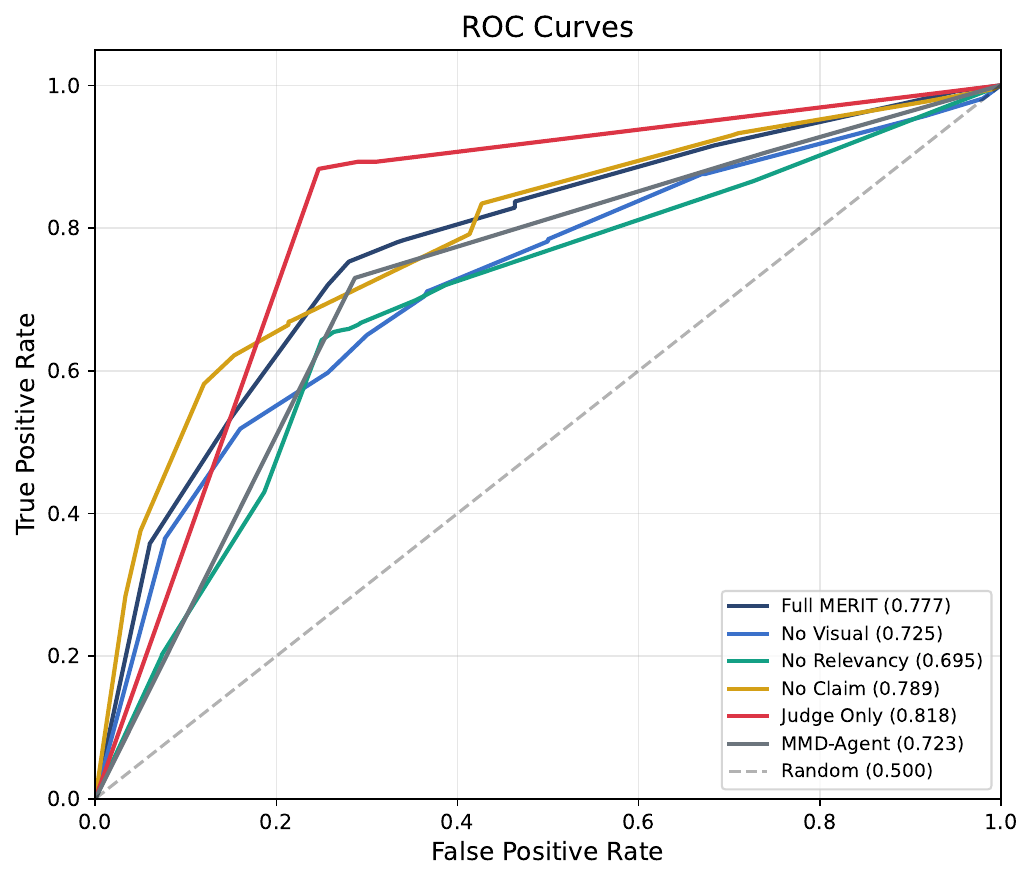}
  \caption{ROC curves on the MMFakeBench validation set. We report ROC-AUC for completeness, although PR-AUC is more informative under class imbalance.}
  \label{fig:roc_curve}
\end{figure}

\end{document}